\documentclass[letterpaper, 10 pt, conference]{ieeeconf}

\IEEEoverridecommandlockouts                              

\usepackage{graphics} \usepackage{epsfig} \usepackage{algorithm}
\usepackage{algpseudocode}
\usepackage{amsmath, amssymb}
\usepackage{xcolor}
\usepackage{graphicx}
\usepackage{subcaption} \usepackage{algorithm}
\usepackage{algpseudocode}
\usepackage{placeins}
\usepackage{balance}
\usepackage{url}

\title{\LARGE \bf
Evidence-Based Landing Site Selection and Vison-Based Landing for UAVs in Unstructured Environments
}

\author{
Sina Sajjadi$^{1}$,
Jacopo Panerati$^{1}$,
Sina Soleymanpour$^{1}$,
Varunkumar Mehta$^{1}$,\\
Farrokh Janabi-Sharifi$^{2}$,
and Iraj Mantegh$^{1}$\thanks{This work was supported by the National Research Council Canada through the Artificial Intelligence for Logistics (AI4L) and Integrated Air Mobility (IAM) programs.}\thanks{$^{1}$S. Sajjadi, J. Panerati, S. Soleymanpour, V. Mehta, and I. Mantegh are with the Aerospace Research Centre, National Research Council Canada, Montreal, QC, Canada
{\tt\small \{sina.sajjadi, jacopo.panerati, sina.soleymanpour, varunkumar.mehta, iraj.mantegh\}@cnrc-nrc.gc.ca}}\thanks{$^{2}$F. Janabi-Sharifi is with the Department of Mechanical and Industrial Engineering, Toronto Metropolitan University, Toronto, ON, Canada
{\tt\small fsharifi@torontomu.ca}}}
 
\begin{document}

\maketitle
\thispagestyle{empty}
\pagestyle{empty}

\begin{abstract}
Autonomous landing in cluttered or unstructured environments remains a safety-critical challenge for unmanned aerial vehicles (UAVs), particularly under noisy perception caused by sensor uncertainty and platform-induced disturbances such as vibration. This paper presents an evidence-based probabilistic framework for autonomous UAV landing that explicitly separates decision-making under uncertainty from execution via visual servoing. Landing safety is modeled as a latent variable and inferred through recursive accumulation of frame-wise visual likelihoods derived from flatness, slope, and obstacle cues, yielding a temporally consistent belief map that is robust to transient perception errors. Physical feasibility is enforced through a hard geometric constraint based on the minimum required landing radius of the UAV, ensuring that undersized but visually appealing regions are rejected. The final landing site is selected using constrained maximum a posteriori estimation. Once selected, the UAV locks onto the target region using ORB feature tracking and performs precise alignment and descent via image-based visual servoing (IBVS). The proposed approach is validated through both real-world laboratory experiments and high-fidelity simulations in Nvidia Isaac Sim, demonstrating consistent, cautious, and stable landing behavior across domains.

\end{abstract}

\section{Introduction}

\begin{figure}\centering
\includegraphics[width=1.0\columnwidth]{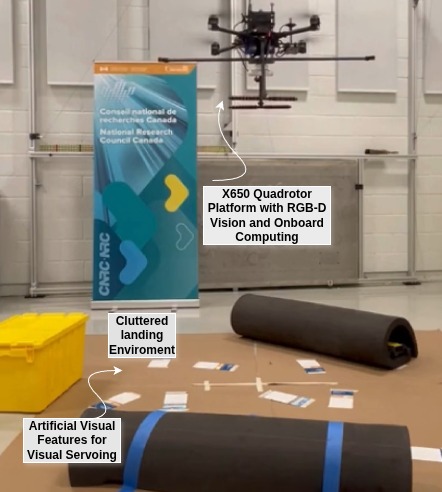}
\caption{Experimental setup used for laboratory validation. The X650 quadrotor carries a downward-facing depth camera and onboard computing to (i) estimate per-frame geometric likelihoods, (ii) accumulate temporal evidence into a belief map for landing-site selection, and (iii) execute the landing using ORB-based tracking and image-based visual servoing.}
\label{fig:frontpage}
\end{figure}

Autonomous landing is a fundamental capability for uncrewed aircraft systems (UAS), defined as an uncrewed aerial vehicle together with its onboard sensors, computation, and control infrastructure. Safe and reliable landing directly impacts mission success, autonomy, and operational safety. While controlled landing conditions can be ensured in structured environments, many emerging UAS applications such as inspection, logistics, emergency response, and indoor or confined operations require landing in cluttered, partially unknown, or dynamically changing surroundings. In these settings, the uncrewed aircraft must autonomously identify a landing site that is not only geometrically suitable but also sufficiently safe under uncertain and noisy sensor observations.

Reliable landing site selection remains challenging due to the multiple sources of uncertainty inherent to onboard perception. Vision based sensing using monocular or depth cameras is affected by measurement noise, limited resolution, partial occlusions, and surface reflectance variations. These challenges are further exacerbated by platform induced disturbances such as vibration and motion blur, which introduce transient artifacts in the perceived terrain geometry. As a result, regions that appear flat and obstacle free in individual frames may in fact be too small, sloped, or unsafe for landing when evaluated over time.

Conventional landing strategies for UAS often rely on instantaneous geometric metrics or frame wise scoring of candidate regions, implicitly assuming that momentary perceptual cues accurately reflect landing safety. However, in safety critical autonomous systems, landing decisions must account not only for instantaneous appearance but also for uncertainty, persistence of evidence, and physical feasibility constraints dictated by the aircraft landing footprint. These challenges motivate the need for landing frameworks that explicitly reason about uncertainty over time, enforce physical feasibility, and decouple high level decision making from low level execution.

This paper presents an evidence-based probabilistic framework for autonomous UAV landing in cluttered environments. Landing safety is modeled as a latent variable and inferred through recursive temporal accumulation of geometric visual cues, explicitly accounting for perception uncertainty and transient sensing artifacts. A hard geometric feasibility constraint based on the vehicle landing footprint is enforced to prevent selection of visually favorable but physically undersized regions. High-level landing site selection is decoupled from execution by transitioning to feature-based image-based visual servoing for stable terminal alignment and descent. The proposed approach is validated through both high-fidelity simulation and real-world laboratory experiments on a full-scale multirotor platform.

\section{Related Work}
\label{sec:related}

Autonomous landing for uncrewed aircraft systems has been the subject of extensive research over the past decade. A recent and comprehensive survey by Semerikov \textit{et al.}~\cite{Semerikov2025Survey} reviews vision based autonomous landing methods published between 2018 and 2025 and categorizes existing approaches into geometric, learning based, and hybrid strategies. The survey identifies persistent challenges related to perception uncertainty, lack of temporal reasoning, and limited enforcement of physical feasibility constraints, particularly in cluttered or unstructured environments.

Early approaches to autonomous landing rely primarily on geometric reasoning over reconstructed terrain. Common techniques include planar fitting, slope estimation, surface roughness analysis, and obstacle proximity evaluation using monocular, stereo, or RGB-D sensing~\cite{Scherer2012,Johnson2015,Bosch2014}. These methods are attractive due to their interpretability and low computational cost, which makes them suitable for onboard implementation. However, they typically operate on instantaneous observations and employ fixed thresholds, implicitly assuming that frame level geometric cues accurately reflect landing safety. As noted in~\cite{Semerikov2025Survey}, this assumption is fragile in the presence of sensor noise, partial observability, and platform induced disturbances such as vibration.

To improve robustness, several works formulate landing site selection as a cost map or scoring problem, where candidate regions are evaluated using weighted combinations of geometric and appearance based features~\cite{Scherer2012,Barry2015,Hinzmann2018}. While cost-based formulations provide a unified representation of landing suitability, the choice of weights is often heuristic and environment dependent. Moreover, most implementations remain frame based and do not explicitly model uncertainty or temporal persistence. As a result, transient regions or geometrically insufficient flat patches may be favored momentarily, leading to unstable or overly optimistic landing decisions.

More recent research has explored learning-based approaches for landing site detection and classification. Convolutional neural networks, semantic segmentation, and graph-based representations have been applied to identify suitable landing areas from images or point clouds~\cite{Zhang2018,Kim2021,Planche2022,Chen2025TASE}. These approaches demonstrate improved performance in structured datasets and simulation environments. However, they typically rely on large labeled training sets and produce point estimates without explicit representation of epistemic uncertainty. Their performance may degrade under domain shift, sensor noise, or environmental conditions not represented during training, which limits their applicability in safety-critical landing scenarios~\cite{Semerikov2025Survey}.

Beyond classical learning-based formulations, recent studies illustrate the broader and sustained research interest in autonomous UAV landing under complex and constrained operational conditions. Vision-based guidance laws have been proposed for landing on moving platforms and dynamic targets, highlighting challenges in perception-driven terminal descent and real-time alignment~\cite{Jang2025VisionOptimal}. Depth-based approaches have also been investigated for safe landing in GPS-denied environments, relying on onboard geometric reasoning from RGB-D sensing~\cite{Cerda2025DepthSafe}. In parallel, data-driven decision-making strategies, including reinforcement learning and semantic perception, continue to be explored for landing site assessment and control using monocular or depth imagery~\cite{Houichime2025RLMonocular,Baidya2024UAVLanding}. While these methods report promising results in specific scenarios, they typically operate on frame-wise estimates and do not explicitly model temporal uncertainty or enforce hard physical feasibility constraints dictated by the vehicle landing footprint, motivating the evidence-based formulation adopted in this work.

Several works have also investigated visual servoing techniques for precise landing and alignment of uncrewed aircraft. Image-based and pose-based visual servoing formulations have been shown to provide reliable convergence properties once a landing target is defined~\cite{Chaumette2006,Corke2011,Kendoul2012}. These methods are well suited for terminal landing control but generally assume that a suitable landing site has already been selected. As such, they do not address the upstream decision-making problem under perceptual uncertainty and are often tightly coupled with the detection process~\cite{Barry2015}.

Despite these advances, relatively few works explicitly formulate landing site selection as a probabilistic inference problem that accumulates evidence over time while enforcing hard physical feasibility constraints dictated by the UAS landing footprint~\cite{Kurdel2024Aerospace}. In particular, the integration of temporal belief accumulation, geometric feasibility filtering, and decoupled visual servoing execution within a unified framework remains underexplored. This gap motivates the approach presented in this work.

Decision making under uncertainty is decoupled from low level execution by transitioning to feature based tracking using ORB features~\cite{Rublee2011ORB} once a landing site is selected, followed by precise alignment and descent using image based visual servoing.

The remainder of this paper is organized as follows. Section~\ref{sec:formulation} presents the problem formulation and introduces the probabilistic landing site selection framework, including the latent safety model, likelihood construction, belief accumulation, and geometric feasibility constraint. The section also describes the image based tracking and visual servoing strategy used for landing execution once a feasible site is selected. Section~\ref{sec:results} presents experimental results obtained in a high fidelity simulation environment using NVIDIA Isaac Sim~\cite{NVIDIAIsaacSim}, followed by real world laboratory experiments that validate the proposed approach under sensor noise and platform induced disturbances. Finally, Section~\ref{sec:conclusions} concludes the paper and discusses directions for future work.

\section{Problem Formulation}
\label{sec:formulation}

We address autonomous landing for an uncrewed aircraft system (UAS) operating in cluttered or unstructured environments using onboard depth sensing. At discrete time steps $t \in \mathbb{N}$, the UAS acquires a depth image $D_t \in \mathbb{R}^{H \times W}$ from an onboard RGB-D sensor (Intel RealSense). Let $\Omega \subset \mathbb{Z}^2$ denote the image grid. Each depth measurement is modeled as
\begin{equation}
D_t(\mathbf{u}) = d_t(\mathbf{u}) + \varepsilon_t(\mathbf{u}), \quad \mathbf{u} \in \Omega,
\end{equation}
where $d_t(\mathbf{u})$ is the true range and $\varepsilon_t(\mathbf{u})$ captures range measurement error, missing returns, and artifacts induced by motion blur and platform vibration. Since instantaneous geometric estimates derived from $D_t$ can be unreliable, landing site selection is formulated as probabilistic inference with temporal evidence accumulation.

\subsection{Candidate Regions and Observations}

From each depth image $D_t$, a set of candidate landing regions
\begin{equation}
\mathcal{R}_t = \{r_{t,1}, \dots, r_{t,N_t}\}
\end{equation}
is extracted, where each region $r_{t,i} \subset \Omega$ is a contiguous set of pixels that satisfies basic geometric screening criteria. Similar region-based representations for landing site analysis have been widely adopted in prior work~\cite{Scherer2012,Bosch2014,Hinzmann2018}.

For each region $r_{t,i}$, a feature vector
\begin{equation}
\mathbf{y}_{t,i} = \phi(D_t, r_{t,i}) \in \mathbb{R}^{m}
\end{equation}
is computed, where $\phi(\cdot)$ extracts depth-derived geometric cues, including surface flatness, surface slope, and obstacle proximity. These cues have been shown to be effective indicators of landing suitability in both deterministic and cost-based formulations~\cite{Johnson2015,Barry2015}.

\subsection{Latent Safety State}

Each region $r_{t,i}$ is associated with a latent binary safety variable
\begin{equation}
S_{t,i} \in \{0,1\}, \quad S_{t,i} = 1 \text{ denotes safe for landing}.
\end{equation}
The safety state is not directly observable and must be inferred from noisy geometric observations. Consistent with probabilistic terrain assessment approaches~\cite{Kurdel2024Aerospace}, we assume a conditionally independent observation model given safety:
\begin{equation}
p(\mathbf{y}_{t,i} \mid S_{t,i}) = p(\phi(D_t,r_{t,i}) \mid S_{t,i}).
\end{equation}

\subsection{Likelihood Construction from Geometric Cues}

For each candidate region, three depth-derived cues are considered. Flatness is estimated by fitting a local plane to the depth measurements within $r_{t,i}$ and computing the normalized residual error $f_{t,i}$. Surface slope is obtained from the angle between the estimated plane normal and the gravity-aligned vertical axis, yielding slope magnitude $s_{t,i}$. Obstacle proximity is quantified using depth discontinuities or height variations within or near the region, producing an obstacle score $o_{t,i}$. Similar geometric measures are commonly used in vision-based landing systems~\cite{Bosch2014,Johnson2015}.

Each cue is mapped to a likelihood term conditioned on the safe landing hypothesis:
\[
\ell_f(f_{t,i}), \quad \ell_s(s_{t,i}), \quad \ell_o(o_{t,i}),
\]
where $\ell(\cdot)$ is a bounded, monotonically decreasing function that assigns higher likelihood to geometrically favorable values. Assuming conditional independence between cues, the overall likelihood is constructed as
\begin{equation}
p(\mathbf{y}_{t,i} \mid S_{t,i}=1) \propto
\ell_f(f_{t,i})^{w_f}
\ell_s(s_{t,i})^{w_s}
\ell_o(o_{t,i})^{w_o},
\end{equation}
where $w_f$, $w_s$, and $w_o$ are nonnegative weights encoding the relative importance of flatness, slope, and obstacle clearance. This weighted formulation follows common practice in cost-map and probabilistic landing approaches while preserving interpretability~\cite{Scherer2012,Hinzmann2018}. For notational convenience, the likelihood under the safe-landing hypothesis
is denoted by
\begin{equation}
L^1_{t,i} = p(\mathbf{y}_{t,i} \mid S_{t,i}=1).
\end{equation}
Similarly, the likelihood under the unsafe hypothesis is denoted by
\begin{equation}
L^0_{t,i} = p(\mathbf{y}_{t,i} \mid S_{t,i}=0),
\end{equation}
and is defined analogously by assigning higher probability to regions that
violate flatness, slope, or obstacle clearance criteria.

\subsection{Temporal Persistence and Belief Update}

Landing safety is expected to be temporally persistent for static environments, while perception noise may cause short-lived fluctuations in geometric cues. This behavior is modeled using a first-order Markov process:
\begin{equation}
p(S_{t,i} \mid S_{t-1,i}) =
\begin{cases}
\alpha, & S_{t,i} = S_{t-1,i}, \\
1-\alpha, & S_{t,i} \neq S_{t-1,i},
\end{cases}
\quad \alpha \in (0.5,1).
\end{equation}

The belief that region $r_{t,i}$ is safe at time $t$ is defined as
\begin{equation}
b_{t,i} = p(S_{t,i}=1 \mid \mathbf{y}_{1:t,i}),
\end{equation}
and is updated recursively using Bayesian filtering:
\begin{equation}
\bar{b}_{t,i} = \alpha b_{t-1,i} + (1-\alpha)(1-b_{t-1,i}),
\end{equation}
\begin{equation}
b_{t,i} =
\frac{p(\mathbf{y}_{t,i} \mid S_{t,i}=1)\,\bar{b}_{t,i}}
{p(\mathbf{y}_{t,i} \mid S_{t,i}=1)\,\bar{b}_{t,i}
+ p(\mathbf{y}_{t,i} \mid S_{t,i}=0)\,(1-\bar{b}_{t,i})}.
\end{equation}
This recursion suppresses transient false positives caused by noisy depth measurements and platform-induced disturbances.

\subsection{Hard Geometric Feasibility Constraint and Site Selection}

While the posterior belief $b_{t,i}$ captures probabilistic landing safety under perceptual uncertainty, it does not guarantee physical feasibility. A region may exhibit high geometric likelihood yet be physically undersized relative to the vehicle landing footprint. To enforce safety at the geometric level, a hard feasibility constraint is introduced.

Let $\rho(r_{t,i})$ denote the maximum inscribed radius of region $r_{t,i}$ projected onto the ground plane, defined as the radius of the largest Euclidean disk fully contained within the region boundary. In practice, this quantity is computed from the binary region mask using a Euclidean distance transform and selecting the maximum interior distance to the boundary.

A candidate region is considered geometrically feasible if
\begin{equation}
\mathcal{F}(r_{t,i}) = \mathbb{I}\!\left[\rho(r_{t,i}) \ge \rho_{\min}\right],
\end{equation}
where $\mathbb{I}[\cdot]$ denotes the indicator function, returning $1$ if the condition is satisfied and $0$ otherwise, and $\rho_{\min}$ is the minimum required landing radius determined by the vehicle footprint and safety margin.

At decision time $t^\star$, the landing site is selected via constrained maximum a posteriori (MAP) estimation over the feasible candidate set:
\begin{equation}
r^\star
=
\arg\max_{r_{t^\star,i} \in \mathcal{R}_{t^\star}^{\mathrm{feas}}}
b_{t^\star,i},
\end{equation}
where
\begin{equation}
\mathcal{R}_{t^\star}^{\mathrm{feas}}
=
\left\{
r_{t^\star,i} \in \mathcal{R}_{t^\star}
\;\middle|\;
\mathcal{F}(r_{t^\star,i}) = 1
\right\}.
\end{equation}

The operator $\arg\max$ returns the feasible region attaining the highest posterior belief. This constrained formulation ensures that landing decisions are both probabilistically justified and physically admissible, preventing visually favorable but geometrically insufficient regions from being selected.

\subsection{Execution via Feature Tracking and Visual Servoing}

Once a feasible landing site $r^\star$ is selected, high-level decision making under uncertainty is decoupled from low-level execution. At this stage, the objective is no longer to reason about landing safety, but to accurately align the vehicle with the selected region and execute a stable vertical descent. This is achieved through feature-based image-based visual servoing (IBVS), using ORB features~\cite{Rublee2011ORB} extracted within the selected region.

\subsubsection{Feature Representation}

Let $\mathcal{F}_t = \{\mathbf{s}_{t,j}\}_{j=1}^{N_t}$ denote the set of $N_t$ ORB feature points successfully tracked at time $t$ within the selected landing region, where each feature
\[
\mathbf{s}_{t,j} = [u_{t,j}, v_{t,j}]^\top
\]
corresponds to pixel coordinates in the image plane. Rather than controlling each feature independently, a single representative visual feature is constructed as the centroid of the tracked feature set:
\begin{equation}
\mathbf{s}_t = \frac{1}{N_t}\sum_{j=1}^{N_t} \mathbf{s}_{t,j}.
\end{equation}
This centroid provides a robust aggregate representation of the landing region that is resilient to individual feature loss and tracking noise, as observed in both simulation and laboratory experiments.

The desired feature location is defined as
\[
\mathbf{s}^\star = [u^\star, v^\star]^\top,
\]
corresponding to the image center, which represents perfect alignment between the camera optical axis and the selected landing site.

The image-space error is therefore
\begin{equation}
\mathbf{e}_t = \mathbf{s}_t - \mathbf{s}^\star.
\end{equation}

\subsubsection{Interaction Matrix Approximation}

Assuming that the tracked features lie on a locally planar surface and share a common depth $Z_t$ relative to the camera, the time variation of the centroid feature can be approximated by the standard IBVS relationship:
\begin{equation}
\dot{\mathbf{s}}_t = \mathbf{L}(\mathbf{s}_t, Z_t)\,\mathbf{v}_c,
\end{equation}
where $\mathbf{v}_c = [v_x, v_y, v_z]^\top$ denotes the camera translational velocity expressed in the camera frame, and $\mathbf{L}(\mathbf{s}_t, Z_t)$ is the interaction matrix given by
\begin{equation}
\mathbf{L}(\mathbf{s}_t, Z_t) =
\begin{bmatrix}
-\frac{1}{Z_t} & 0 & \frac{u_t}{Z_t} \\
0 & -\frac{1}{Z_t} & \frac{v_t}{Z_t}
\end{bmatrix}.
\end{equation}

The depth $Z_t$ is obtained from the onboard RGB-D sensor by averaging depth measurements within the selected region. Rotational motion is neglected during the terminal landing phase, as yaw is either held constant or regulated independently by the flight controller. This assumption is consistent with the experimental setup and simplifies the control law while remaining effective in practice.

\subsubsection{Visual Servo Control Law}

A stabilizing IBVS control law is defined as
\begin{equation}
\mathbf{v}_c = -\lambda\,\mathbf{L}(\mathbf{s}_t, Z_t)^{+}\,\mathbf{e}_t,
\end{equation}
where $\lambda > 0$ is a scalar control gain and $(\cdot)^{+}$ denotes the Moore--Penrose pseudoinverse. The resulting translational velocity command directly regulates the horizontal alignment of the vehicle with the landing site through $(v_x, v_y)$, while the vertical component $v_z$ governs the descent rate.

In practice, this formulation produces smooth, bounded velocity commands and ensures monotonic reduction of the image-space error, as confirmed by the experimental results presented in Section~III. The use of an aggregated centroid feature, rather than individual feature-level control, further improves robustness to intermittent feature loss and sensor noise during the terminal landing phase.

\begin{algorithm}[t]
\caption{Evidence-Based Probabilistic Landing Site Selection and Execution}
\label{alg:ebpls}
\begin{algorithmic}[1]
\Require Depth stream $\{D_t\}$, minimum landing radius $\rho_{\min}$, initial belief $b_{0,i}$, persistence $\alpha$, cue weights $\{w_f,w_s,w_o\}$, belief threshold $\tau$, IBVS gain $\lambda$
\Ensure Selected landing site $r^\star$ and executed landing

\State Initialize beliefs $\{b_{0,i}\}$; $t \gets 1$; $r^\star \gets \emptyset$
\While{$r^\star = \emptyset$}
    \State Acquire depth image $D_t$
    \State Extract candidate regions $\mathcal{R}_t = \{r_{t,i}\}$ from $D_t$
    \State Associate regions across time to maintain identities
    \For{each region identity $i$}
        \State Compute geometric cues $(f_{t,i}, s_{t,i}, o_{t,i}) \gets \phi(D_t, r_{t,i})$
        \State Compute $L^1_{t,i} = p(\mathbf{y}_{t,i} \mid S_{t,i}=1)$
        \State Compute $L^0_{t,i} = p(\mathbf{y}_{t,i} \mid S_{t,i}=0)$
        \State Predict belief $\bar b_{t,i} \gets \alpha b_{t-1,i} + (1-\alpha)(1-b_{t-1,i})$
        \State Update belief
        \[
        b_{t,i} \gets 
        \frac{L^1_{t,i}\bar b_{t,i}}
        {L^1_{t,i}\bar b_{t,i} + L^0_{t,i}(1-\bar b_{t,i})}
        \]
        \State Enforce feasibility $\mathcal{F}(r_{t,i}) \gets \mathbb{I}[\rho(r_{t,i}) \ge \rho_{\min}]$
    \EndFor
    \State $\mathcal{R}^{\mathrm{feas}}_t \gets \{r_{t,i} \in \mathcal{R}_t \mid \mathcal{F}(r_{t,i})=1\}$
    \If{$\mathcal{R}^{\mathrm{feas}}_t \neq \emptyset$}
        \State $i^\star \gets \arg\max_{r_{t,i}\in\mathcal{R}^{\mathrm{feas}}_t} b_{t,i}$
        \If{$b_{t,i^\star} \ge \tau$}
            \State Select landing site $r^\star \gets r_{t,i^\star}$
            \State Initialize ORB feature tracking in $r^\star$
        \EndIf
    \EndIf
    \State $t \gets t+1$
\EndWhile

\While{not landed}
    \State Track ORB features and compute centroid $\mathbf{s}_t$
    \State Compute image error $\mathbf{e}_t \gets \mathbf{s}_t - \mathbf{s}^\star$
    \State Compute velocity command $\mathbf{v}_c \gets -\lambda\,\mathbf{L}(\mathbf{s}_t,Z_t)^{+}\mathbf{e}_t$
    \State Send velocity setpoint to the flight controller
\EndWhile
\end{algorithmic}
\end{algorithm}

\begin{table}[t]
\caption{Experimental parameters and symbols used in the landing framework.}
\label{tab:parameters}
\centering
\begin{tabular}{l c c}
\hline
\textbf{Description} & \textbf{Symbol} & \textbf{Value} \\
\hline
Belief update rate            & $f_s$                & 10 Hz \\
Flatness weight               & $w_f$                & 0.4 \\
Slope weight                  & $w_s$                & 0.2 \\
Obstacle proximity weight     & $w_o$                & 0.4 \\
Temporal persistence          & $\alpha$             & 0.95 \\
Initial belief                & $b_{0,i}$            & 0.5 \\
Belief threshold              & $\tau$               & 0.75 \\
Minimum landing radius        & $\rho_{\min}$        & 0.55 m \\
IBVS control gain             & $\lambda$            & 0.8 \\
Maximum lateral velocity      & $\|\mathbf{v}\|_{\max}$ & 0.25 m/s \\
Maximum vertical descent rate & $|v_z|_{\max}$       & 0.30 m/s \\
\hline
\end{tabular}
\end{table}

\section{Results}
\label{sec:results}

\begin{figure*}\centering
\includegraphics[width=0.9\textwidth]{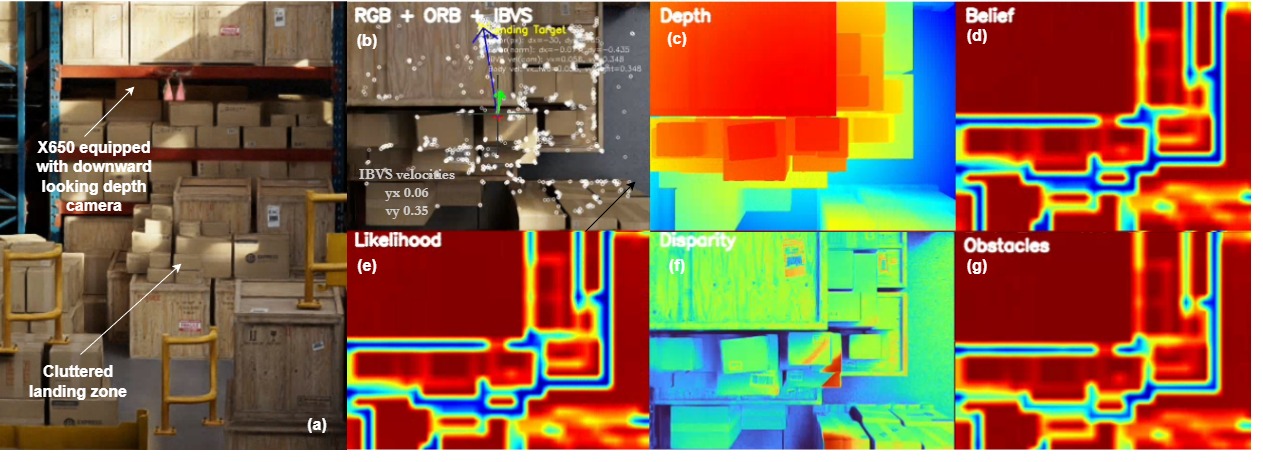}
\caption{Simulation snapshot of the proposed landing pipeline in a cluttered warehouse scene (NVIDIA Isaac Sim). (a) RGB view of the landing area. (b) RGB overlay showing ORB features and the IBVS target/centroid used during terminal alignment. (c) Depth map. (d) Accumulated belief map (posterior safety) after temporal evidence integration. (e) Frame-wise likelihood map computed from geometric cues. (f) Disparity representation (depth structure) used by the cue extraction. (g) Obstacle-related map highlighting clutter boundaries and unsafe regions. In the simulation scenario, the hard constraint on landing site size was not enforced.} 
\label{fig:sim_overview}
\end{figure*}

\begin{figure*}{\centering
\includegraphics[width=1.04\textwidth,
trim={0.8cm 0.0cm 0.0cm 0.0cm},clip
                ]{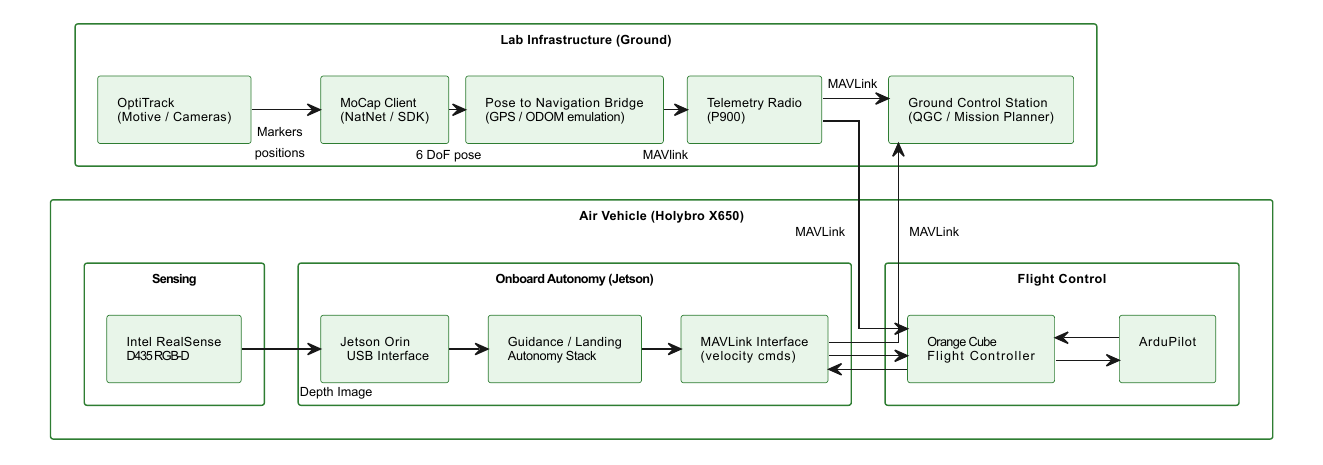}
}
\caption{Laboratory experimental architecture illustrating OptiTrack-based indoor positioning, onboard RGB-D perception, MAVLink-enabled guidance, and vision-guided landing on a Holybro X650 platform. The Jetson autonomy stack performs landing-site inference and IBVS-based terminal control, while ArduPilot executes low-level stabilization.}
\label{fig:lab_architecture}
\end{figure*}

\begin{figure*}\centering
\includegraphics[width=0.8\textwidth]{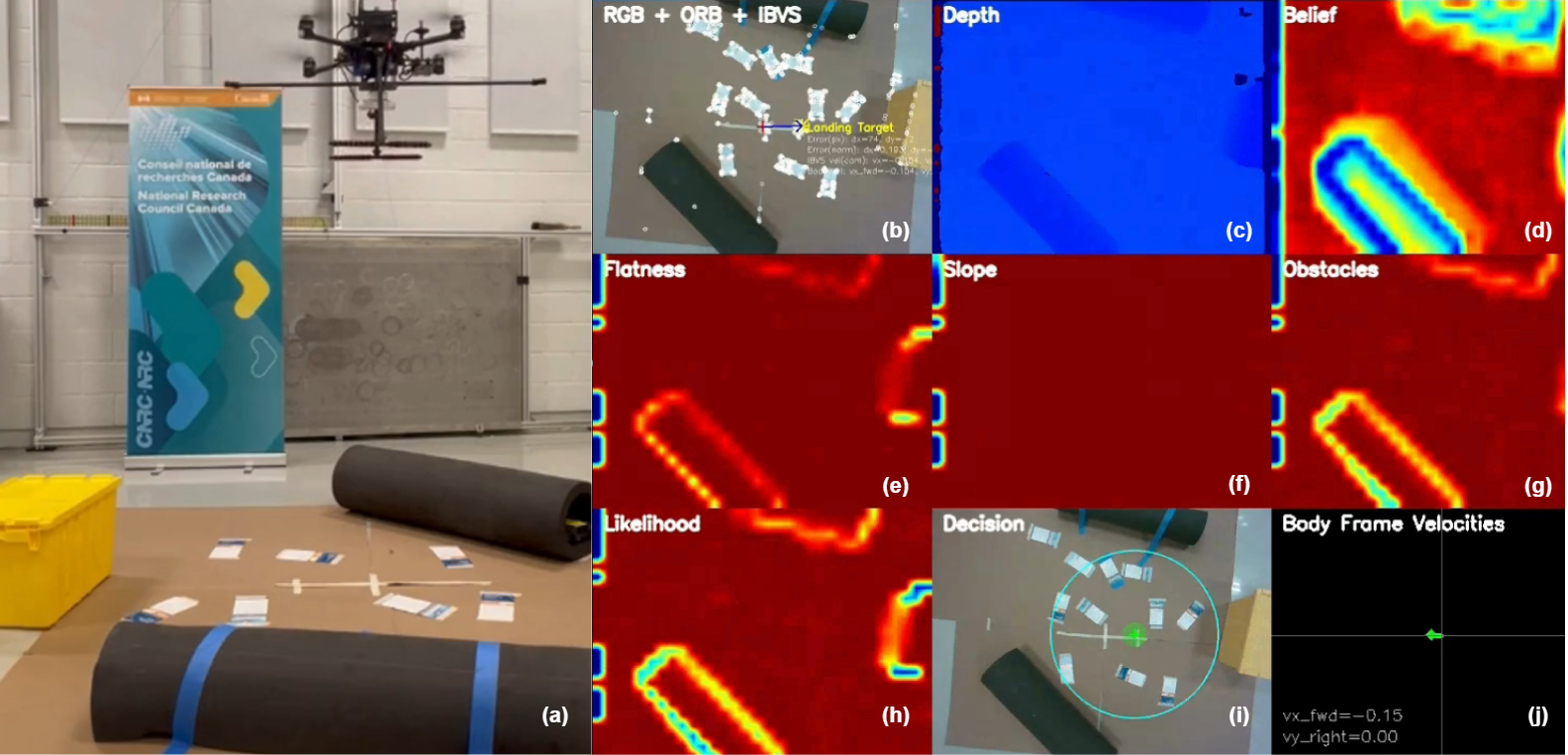}
\caption{Laboratory snapshot of belief formation and terminal execution. (a) RGB view of the landing scene. (b) ORB feature tracks and IBVS target/centroid overlay used for alignment. (c) Depth map. (d) Accumulated belief map (posterior safety). (e--g) Cue-specific responses (flatness, slope, and obstacle proximity). (h) Combined frame-wise likelihood. (i) Body-frame velocity command visualization during descent (lateral, longitudinal, and vertical components).}
\label{fig:lab_overview}
\end{figure*}

\begin{figure*}\centering
\includegraphics[width=0.8\textwidth]{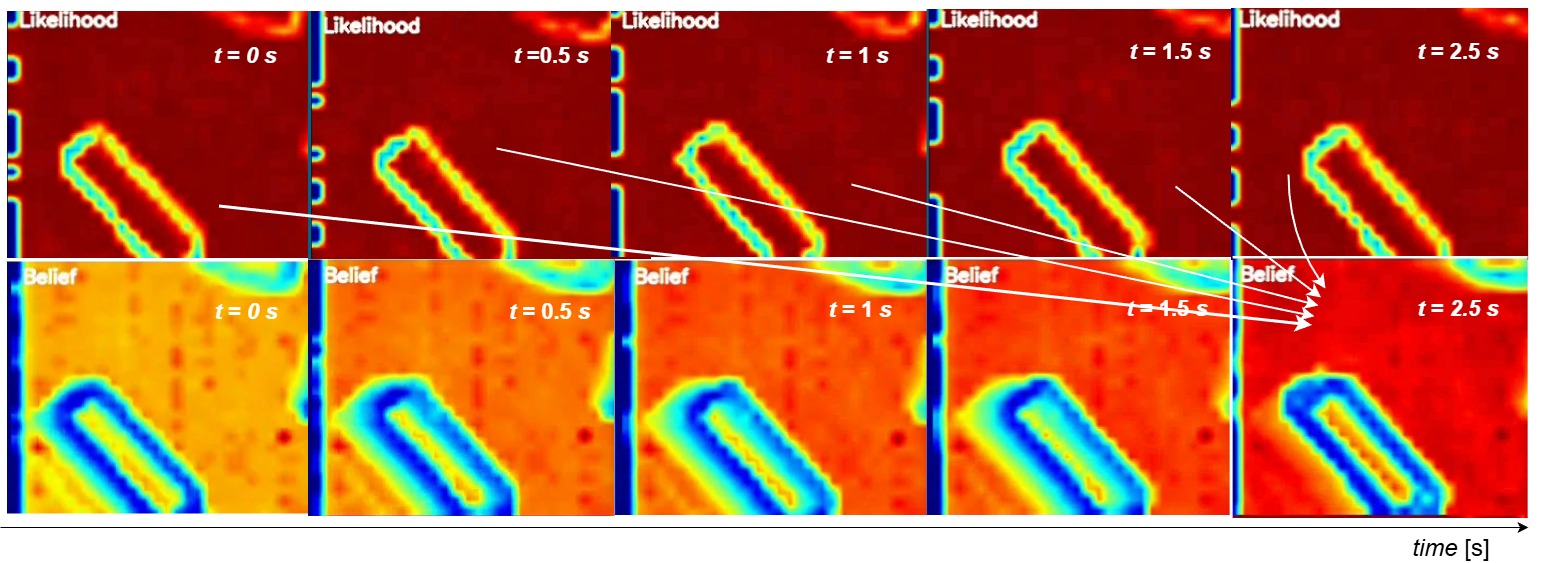}
\caption{Temporal evolution of instantaneous likelihood (top row) and accumulated belief (bottom row) over time (with example timestamps). The likelihood can fluctuate due to depth noise and view-dependent artifacts, while the belief remains temporally consistent by integrating evidence and suppressing transient false positives.}
\label{fig:belief_evolution}
\end{figure*}

This section reports results from high-fidelity simulation and laboratory experiments. The objectives are to \emph{(i)} verify that temporal belief accumulation suppresses transient perception artifacts, \emph{(ii)} confirm that the ``landing-footprint feasibility'' hard constraint prevents the selection of visually appealing but undersized regions, and \emph{(iii)} demonstrate stable terminal execution via feature tracking and IBVS, once a site is selected.

\subsection{Simulation Results in NVIDIA Isaac Sim}

Simulation experiments were carried out with NVIDIA Isaac Sim~\cite{NVIDIAIsaacSim} in a cluttered warehouse environment containing shelves, boxes, and partially planar surfaces representative of unstructured indoor landing scenarios. The simulated vehicle was equipped with a downward-facing RGB-D sensing suite, implemented using Isaac Sim Replicator cameras to generate synchronized RGB images and depth measurements. The perception pipeline computes per-region geometric cues (flatness, slope, obstacle proximity) from the depth stream and converts them into frame-wise likelihood maps, which are then integrated over time using the Bayesian belief update described in Section~\ref{sec:formulation}.

Fig.~\ref{fig:sim_overview} provides a representative snapshot of the full perception pipeline in simulation. The RGB view of the scene is shown in Fig.~\ref{fig:sim_overview}(a). The system extracts and tracks ORB features to support terminal alignment (Fig.~\ref{fig:sim_overview}(b)), while depth and derived maps drive the probabilistic site assessment. Specifically, the depth/disparity representations (Fig.~\ref{fig:sim_overview}(c,f)) enable computation of obstacle and terrain structure; the resulting frame-wise likelihood map (Fig.~\ref{fig:sim_overview}(e)) may exhibit spurious high responses due to sensor artifacts and depth discontinuities. The belief map (Fig.~\ref{fig:sim_overview}(d)), obtained via temporal evidence accumulation, is smoother and more conservative than the instantaneous likelihood, reflecting the intended suppression of short-lived false positives. The obstacle-related map (Fig.~\ref{fig:sim_overview}(g)) further penalizes regions near depth edges and clutter boundaries, reducing the probability assigned to geometrically risky candidates.

The temporal behavior observed during our experimental evaluation is shown in Fig.~\ref{fig:belief_evolution}, which illustrates the evolution of instantaneous likelihood and accumulated belief over multiple time steps and provides insight into how temporal evidence accumulation stabilizes landing decisions. The simulations were conducted using a custom multirotor dynamics model developed and integrated within NVIDIA Isaac Sim, enabling realistic vehicle motion, sensor feedback, and closed-loop autonomy evaluation. In simulation, frame-wise likelihood maps fluctuate rapidly due to viewpoint changes and depth sensing noise, while the accumulated belief maps remain temporally consistent and assign high confidence only to regions that persistently satisfy the geometric criteria. The hard landing footprint constraint was not enforced in simulation, allowing the probabilistic behavior to be analyzed independently of physical size limitations. Across repeated simulation trials, the maximum a posteriori (MAP) landing site selection, introduced in Section~\ref{sec:formulation}, consistently favored regions with high accumulated belief, providing valuable insight for subsequent laboratory validation where physical feasibility constraints were applied. Overall, the simulation environment served as an effective development and analysis tool to design, test, and refine the vision-based perception and belief update pipeline prior to real-world experiments.

\subsection{Laboratory Experimental Results }

We conducted laboratory experiments to validate the proposed framework under realistic sensing noise, communication latency, and platform-induced disturbances. The experimental platform is a Holybro X650 multirotor with an Orange Cube flight controller running ArduPilot, and an onboard NVIDIA Jetson Orin executing perception, belief updates, and terminal landing control. Indoor global positioning is emulated using an OptiTrack motion capture system.\footnote{A video of the experimental setup and trials is available at: \url{https://youtu.be/T9wAI17ULsg}} 

The experimental architecture is summarized in Fig.~\ref{fig:lab_architecture}. OptiTrack pose estimates are streamed to the flight controller to support stable flight and pre-landing positioning, while the onboard RealSense RGB-D sensor provides the depth stream for landing-site inference. High-level guidance and velocity setpoints from the Jetson autonomy stack are transmitted to ArduPilot via MAVLink, while low-level attitude stabilization and motor control remain onboard the flight controller. This separation allows the landing module to be evaluated as a self-contained onboard autonomy component, while maintaining flight safety through the flight controller’s inner loops.

The parameters listed in Table~\ref{tab:parameters} were used for the laboratory experiments and were not tuned on a per-scene basis. Cue weights were selected to balance surface stability (flatness), physical feasibility (slope), and safety near clutter (obstacle proximity), with obstacle proximity weighted strongly to avoid selecting regions near depth discontinuities. The temporal persistence parameter $\alpha=0.95$ at a belief update rate of 10~Hz enforces conservative temporal smoothing, requiring consistent geometric evidence over multiple frames before committing to a landing decision. The minimum landing radius $\rho_{\min}$ was derived from the experimental vehicle footprint (approximately 0.95~m in diameter) with an added safety margin to account for tracking uncertainty and attitude transients. Velocity limits and the IBVS gain were chosen to ensure smooth, low-speed terminal alignment, consistent with the observed velocity profiles in the experimental results.

A representative landing trial is shown in Fig.~\ref{fig:lab_overview}. The RGB scene (Fig.~\ref{fig:lab_overview}(a)) includes clutter and surfaces with varying geometric suitability. The belief map (Fig.~\ref{fig:lab_overview}(d)) concentrates probability mass on a stable candidate region, while the cue-specific maps (flatness/slope/obstacle) (Fig.~\ref{fig:lab_overview}(e--g)) expose which geometric properties dominate the posterior at that moment. The combined likelihood (Fig.~\ref{fig:lab_overview}(h)) drives the recursive update and, together with the feasibility constraint, determines the selected region. After selection, the system transitions to execution: ORB feature tracking supports estimation of the feature centroid and the IBVS error (Fig.~\ref{fig:lab_overview}(b)), and the resulting body-frame velocity commands (Fig.~\ref{fig:lab_overview}(i)) remain bounded and smooth, consistent with stable closed-loop descent.

Overall, the laboratory results corroborate the simulation findings: \emph{(i)} temporal accumulation mitigates transient sensor artifacts, \emph{(ii)} feasibility filtering prevents undersized regions from being chosen despite locally favorable appearance, and \emph{(iii)} the decoupled execution stage yields stable terminal alignment and descent.

\section{Conclusion and Future Work}
\label{sec:conclusions}

This paper presented an evidence based probabilistic framework for autonomous landing of uncrewed aircraft systems that addresses perception uncertainty while ensuring physical feasibility and execution robustness. Landing safety was modeled as a latent variable and inferred through recursive accumulation of frame wise visual likelihoods, while a hard geometric feasibility constraint ensured compliance with the minimum landing footprint of the aircraft. Landing execution was achieved through feature based tracking and image based visual servoing. Experimental results from high fidelity simulation using NVIDIA Isaac Sim and real world laboratory tests demonstrated stable and conservative landing behavior under sensor noise and platform induced disturbances.

Future work will investigate replacing hand crafted feature tracking with learning based perception modules and incorporating machine learning to adapt belief updates based on environmental context and perception quality. Additional extensions include handling wind disturbances and dynamic obstacles, as well as validation in more complex outdoor environments.

\balance

\section{Acknowledgment}
This work was supported by National Research Council Canada's Integrated Aerial Mobility and AI For Logistics research programs, in collaboration with Toronto Metropolitan University's Robotics, Mechatronics, and Automation Laboratory (RMAL).

\balance

\end{document}